\documentclass[10pt,twocolumn,letterpaper]{article}

\usepackage{wacv}              

\usepackage{graphicx}
\usepackage{amsmath}
\usepackage{amssymb}
\usepackage{booktabs}
\usepackage{multirow}
\newcommand{\tabincell}[2]{\begin{tabular}{@{}#1@{}}#2\end{tabular}}

\usepackage[pagebackref,breaklinks,colorlinks]{hyperref}
\usepackage[accsupp]{axessibility} 

\usepackage[capitalize]{cleveref}
\crefname{section}{Sec.}{Secs.}
\Crefname{section}{Section}{Sections}
\Crefname{table}{Table}{Tables}
\crefname{table}{Tab.}{Tabs.}

\def\l1{\ensuremath{\ell_1}\xspace}
\def\l2{\ensuremath{\ell_2}\xspace}

\newcommand{\vd}{\mathbf{d}}

\newcommand{\vg}{\mathbf{g}}

\makeatletter
\DeclareRobustCommand\onedot{\futurelet\@let@token\@onedot}
\def\@onedot{\ifx\@let@token.\else.\null\fi\xspace}
\def\eg{\emph{e.g}\onedot} 
\def\ie{\emph{i.e}\onedot}

\def\etal{\emph{et al}\onedot}

\makeatother

\usepackage{stmaryrd}
\def\wacvPaperID{336} 
\def\confName{WACV}
\def\confYear{2025}

\begin{document}

\title{Breaking the Frame:\\Visual Place Recognition by Overlap Prediction}
\author{Tong Wei$^1$, Philipp Lindenberger$^2$, Ji{\v{r}}{\'\i} Matas$^1$, Daniel Barath$^{2,3}$\\
$^1$ Visual Recognition Group, FEE, Czech Technical University in Prague\\
$^2$ Computer Vision and Geometry Group, ETH Zurich, $^3$ HUN-REN SZTAKI \\
{\tt\small \{weitong, matas\}@fel.cvut.cz, \{philipp.lindenberger, danielbela.barath\}@inf.ethz.ch}
}
\maketitle

\begin{abstract}
    Visual place recognition methods struggle with occlusion and partial visual overlaps. 
    We propose a novel visual place recognition approach based on overlap prediction, called VOP, shifting from traditional reliance on global image similarities and local features to image overlap prediction.
    VOP proceeds co-visible image sections by obtaining patch-level embeddings using a Vision Transformer backbone and establishing patch-to-patch correspondences without requiring expensive feature detection and matching. 
    Our approach uses a voting mechanism to assess overlap scores for potential database images. It provides a nuanced image retrieval metric in challenging scenarios.
    Experimental results show that VOP leads to more accurate relative pose estimation and localization results on the retrieved image pairs than state-of-the-art baselines on a number of large-scale, real-world indoor and outdoor benchmarks.
    The code is available at \url{https://github.com/weitong8591/vop.git}.
\end{abstract}

\section{Introduction}
\label{sec:intro}

Visual Place Recognition (VPR) is defined as the task of identifying the approximate location where a query image was taken, given a certain tolerance level 
\cite{liu2019stochastic,Torii2018,Doan2019,arandjelovic2016netvlad,Hausler2021,Zaffar2021,Torii2021,Garg2019,Hausler2019,Khaliq2020,Berton2021,Ibrahimi2021,Warburg2020}. 
VPR methods have been widely used in computer vision and robotics~\cite{delhumeau2013revisiting, chen2017only}, such as autonomous, unmanned aerial~\cite{moskalenko2024visual}, terrestrial, and underwater vehicles~\cite{ribeiro2018underwater}, as well as recent AR/VR devices. 

Typically, VPR is approached as an image retrieval problem, where a query image is compared against a large database of posed images and, optionally, a 3D reconstruction of the scene. 
Then, the most similar images retrieved from the database are used to estimate the precise location of the input query image, \eg, via local feature detection and matching.
The complexity of VPR often stems from various challenges including changes in viewpoint, the presence of dynamic objects, illumination differences, occlusion, weather conditions, seasonal 
variations, and large-scale environments, as discussed in the studies by Hong \etal.\cite{hong2019textplace}, Doan \etal.\cite{doan2019scalable}, and Subramaniam \etal.~\cite{subramaniam2018ncc}.

\begin{figure}[t]
    \centering
    \includegraphics[width=1\columnwidth, trim=0mm 0mm 0mm 0mm, clip]{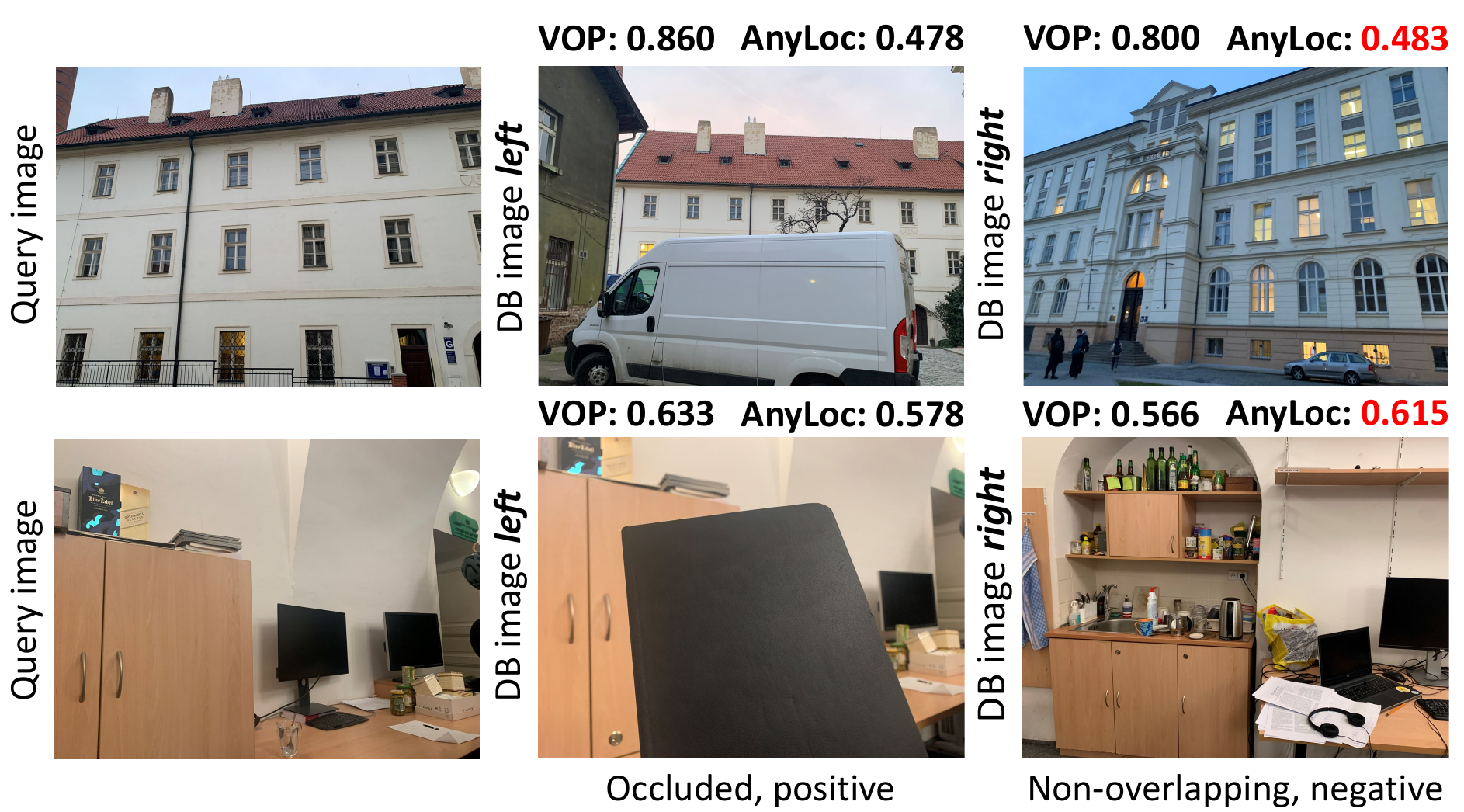}
         \vspace{-1.25em}
     \caption{
     An example where the SOTA AnyLoc~\cite{keetha2023anyloc} scores a negative DB image (\textit{right}, at a different location) higher than an occluded positive example (\textit{left}, the same scene as the query with heavy occlusion).
    {\bf{VOP}} ranks the database (DB) images correctly.}
     \vspace{-0.75em}
    \label{fig:examples}
\end{figure}

Recent approaches implement VPR as an estimation of image similarity and retrieval through learned image embeddings.
Such embeddings are generated by a feature extraction architecture combined with aggregation or pooling mechanisms, such as NetVLAD~\cite{arandjelovic2016netvlad}, CosPlace~\cite{berton2022rethinking}, and AnyLoc~\cite{keetha2023anyloc}. 
In such frameworks, images from the database are retrieved by optimizing the similarity between their embeddings and the query image, \eg, cosine similarity.
One notable drawback of these methods is their sensitivity to partial overlaps caused by occlusion. 
In practice, occlusion can significantly reduce image similarity scores despite the unobstructed portions of the scene, potentially offering valuable cues for localization.
The second drawback lies in training, as they optimize the similarity of the embeddings on positives closer than the negative samples.
This process is inherently difficult because ground truth (GT) data does not provide explicit information about the exact degree of similarity. 
Instead, we build GT patch matches as supervised by 3D reconstruction.
In addition, several reranking-based VPR methods~\cite{Hausler2021, Zhu_2023_R2Former} retrieve the most similar images from a shortlist of database images built by global embeddings.
Such methods improve the accuracy of image retrieval but require more storage. 

Instead of relying on global similarities or local features, we frame the problem as overlap prediction.
This enables us to methodically understand which parts of a particular image are visible without requiring feature detection and matching procedures.
Also, focusing on overlaps makes VOP more robust to occlusion and provides a better understanding of the relations between the query and database images.
VOP encodes individual patches rather than the entire image, thereby facilitating efficient patch-to-patch correspondences. 
These matches are subsequently incorporated into a voting mechanism to calculate an overlap score for potential database images, providing a nuanced metric for retrieval in complex visual environments.

In Fig.~\ref{fig:examples}, we present two retrieval examples using the proposed Visual Overlap Prediction (VOP) method and AnyLoc~\cite{keetha2023anyloc}, a state-of-the-art retrieval solution. 
They process a query image alongside two database images: one sharing the scene with the query but with occlusion, and the other from a different environment. 
AnyLoc incorrectly assigns a higher similarity to the unrelated image. 
In contrast, our proposed approach successfully identifies the correct image and the areas likely to overlap.

\noindent
The contributions of this paper are as follows:
\begin{enumerate}
    \item A novel approach shifting focus from image similarities to
     patch overlap assessment, showing enhanced robustness to challenges, \eg, occlusion. 
    The proposed method, {\bf{VOP}}, 
    learns patch embeddings suitable for visual localization 
    and implements a single radius search across image patches in the database.
    \item The introduction of a robust voting mechanism for image retrieval shows improvements over conventional similarity-based methods with nearest neighbors.
    \item Breaking the frame of looking for close database images in standard VPR methods by considering geometric applications and their evaluation metrics. 
\end{enumerate}

\section{Related Work}

Image retrieval methods for VPR problems can be categorized into two groups, one globally representing the whole image and the other locally for keypoints using handcrafted or learning-based embeddings.

\vspace{1mm}
\noindent
\textbf{Place recognition} is often cast as an image retrieval problem~\cite{GargPlaceRecognition,arandjelovic2016netvlad,Berton2021,Hausler2021,Ibrahimi2021,Kim2017,Peng2021_ICRA,Peng2021_ICCV,Warburg2020,berton2022rethinking,keetha2023anyloc} that consists of two phases. 
In the offline indexing phase, a reference map represented by an image database is created.  
In the online retrieval phase, a query image, taken during a later traverse of the environment, is coarsely localized by identifying its closest matches within the reference map.

To find the closest matches in the indexing structures, traditional methods like Bag of Visual Words~\cite{csurka2004visual} (BoW) identify key local image patches and store them as a codebook. 
BoW is often used in conjunction with 128-dimensional SIFT~\cite{SIFT2004} descriptors both for map creation and image retrieval. 
These high-dimensional features often require approximate nearest neighbor (NN) search or GPUs to accelerate the searching process.
Showcasing the unbroken popularity of such approaches, they are still widely employed in applications, \eg, in ORB-SLAM~\cite{mur2015orb,mur2017orb,campos2021orb}. 

As demonstrated by Zheng \etal.\cite{zheng2017sift}, using CNNs to improve hand-crafted features in various applications, namely through compact (fixed-length) representations, has gradually become the norm. 
Most recent methods utilize learned embeddings generated by a feature extraction backbone with an aggregation or pooling head, \eg, NetVLAD~\cite{arandjelovic2016netvlad}.
In recent works, transformer-based retrieval methods~\cite{el2021training} have been introduced, incorporating metric learning and attention mechanisms. 
These methods utilize local and global features, or a fusion of the two, to identify similar images.

\vspace{1mm}
\noindent
\textbf{Global retrieval methods} usually integrate or combine handcrafted or learning-based local descriptors, such as SIFT~\cite{SIFT2004} through the BOW model~\cite{philbin2007object, sivic2003video}, or employ alternative algorithms~\cite{tolias2016image}. 
There is a growing trend towards learning-based techniques that devise improved methods for feature aggregation. NetVLAD~\cite{arandjelovic2016netvlad} introduces an end-to-end trainable image representation extractor, employing weak supervision through soft assignment to cluster centers. 
Generalized Mean Pooling (GeM)~\cite{radenovic2018fine} proposes a learnable pooling layer that generalizes both max and average pooling, with a CNN serving as the backbone. 
Moreover, AP-GeM~\cite{revaud2019learning} enhances ranking accuracy by minimizing a relaxed global mean Average Precision (mAP) over a set of images. 
A CNN-based solution, CosPlace~\cite{berton2022rethinking}, utilizes the CNN backbone and GeM pooling but approaches  place retrieval as a classification problem. 
MixVPR~\cite{ali2023mixvpr} transitions feature maps into a compact representation space through cascade MLP blocks. 
AnyLoc~\cite{keetha2023anyloc} emerges as a versatile method suitable for various image retrieval contexts, exploring CNN-based or Vision Transformer (ViT)-based foundational models and aggregation techniques to construct a vocabulary for searching in a large database.
More recently, DINOv2-SALAD~\cite{Izquierdo_CVPR_2024_SALAD} finetunes DINOv2 and proposes a new aggregation method based on optimal transport and the relations between features and clusters.
%

\vspace{1mm}
\noindent
\textbf{Image reranking.}
Visual place recognition techniques have increasingly embraced a two-phase approach that pairs global retrieval with subsequent reranking procedure based on local features. 
MultiVLAD~\cite{arandjelovic2013all} is proposed to use the maximum similarity between the query VLAD descriptors and the query as the matching score. 
Similarly, Razavian.~\etal~\cite{sharif2014cnn} rely on the average L2 distance
of each query sub-patch to the reference image.
Patch-NetVLAD~\cite{Hausler2021}, for instance, innovates by integrating NetVLAD with region-level descriptors and multi-scale patch tokens for enhanced geometric verification. 
It effectively merges the strengths of local and global descriptors, offering high resilience to changes in conditions and viewpoints.
RRT~\cite{tan2021instance} and TransVPR~\cite{Wang_2022_TransVPR} further this development by incorporating image-level supervision and attention-based mechanisms, respectively, focusing on spatial relationships and feature relevance within images. 
CVNet and $R^2$Former~\cite{Zhu_2023_R2Former} introduce methodologies for replacing traditional geometric verification with dense feature correlation and transformer-based reranking, focusing on the precise alignment of features and relevance of image pairs.
SuperGlobal~\cite{shao2023global} simplifies the reranking process by relying on global features and K-nearest-neighbors aggregation, showcasing a move towards more scalable and computationally efficient VPR solutions. 
These advancements highlight a trend toward utilizing deep learning techniques and transformers, to refine the accuracy and efficiency of image retrieval.

\section{Visual Overlap Prediction (VOP)}

In this section, we describe the method proposed for efficiently predicting visual overlap in large image collections. 

\subsection{Problem Statement}

%

In place recognition problems, the aim is to find images $I_j\in \mathbb{R}^{H\times W \times C}$ containing the same landmark or scene as the query $I_i\in \mathbb{R}^{H\times W \times C}$.
Usually, it is done by transforming the images to a high-dimensional embedding space and measuring the distances to select the top-$k$ most similar images. 
In this embedding space, an image $I_i$ is represented by a high-dimensional global $\vg_i \in \mathbb{R}^{d_g}$ or local descriptors $\vd_i, \in \mathbb{R}^{n_l \times d_l}$, using encoder $f(\cdot)$, 
where $n_l \in \mathbb{N}$ denotes the number of local features, with the dimensions 
$d_g \in \mathbb{N}$, $d_l \in \mathbb{N}$ .
The matching score, then, can be computed as the similarities of the descriptors, 
$s(I_i, I_j) = sim(\vd_i, \vd_j)$ or $sim(\vg_i, \vg_j)$, for example, as their cosine similarity.

Local features are usually image-dependent and redundant with variable lengths.
Instead, we focus on patch-level representations and redefine the matching score as the overlap between the images. 
It is simple and efficient to achieve patch-level descriptors with less dependency.
Let us split image $I$ into a set $\mathcal{P}$ of rectangular patches by a uniform grid consisting of  $n$ rows and columns.
Given sets of such rectangular patches $\mathcal{P}_i$ and $\mathcal{P}_j$ in the two images, there are multiple ways to define overlap, \eg, a patch-level overlap score as follows:
\begin{equation}
    o(p, q) = \text{volume}(\text{cone}(\mathbf R_i, \mathbf t_i, \mathbf K_i, p), \text{cone}(\mathbf R_j, \mathbf t_j, \mathbf K_j, q)),
\end{equation}
for each pair $(p, q)$, where $p \in \mathcal{P}_i$ and $q \in \mathcal{P}_j$ are the image patches, $\mathbf R \in \text{SO}(3)$ are the absolute rotations of the cameras, $\mathbf t \in \mathbb{R}^3$ are absolute positions, 
$\mathbf K \in \mathbb{R}^{3 \times 3}$ represents the intrinsic camera parameters,
function $\text{cone}(\cdot)$ is the 3D cone defined by the camera pose and the current patch, function $\text{volume}(\cdot, \cdot)$ measures the overlap of two posed cones in three dimensions.
While this measure is easy to calculate given the camera poses and intrinsics, we observed that it does not reflect real-world applications, \eg, 3D reconstruction.
Thus, we reformulate this measure as follows:
\begin{figure}[t]
    \centering
    \includegraphics[width=0.85\columnwidth]{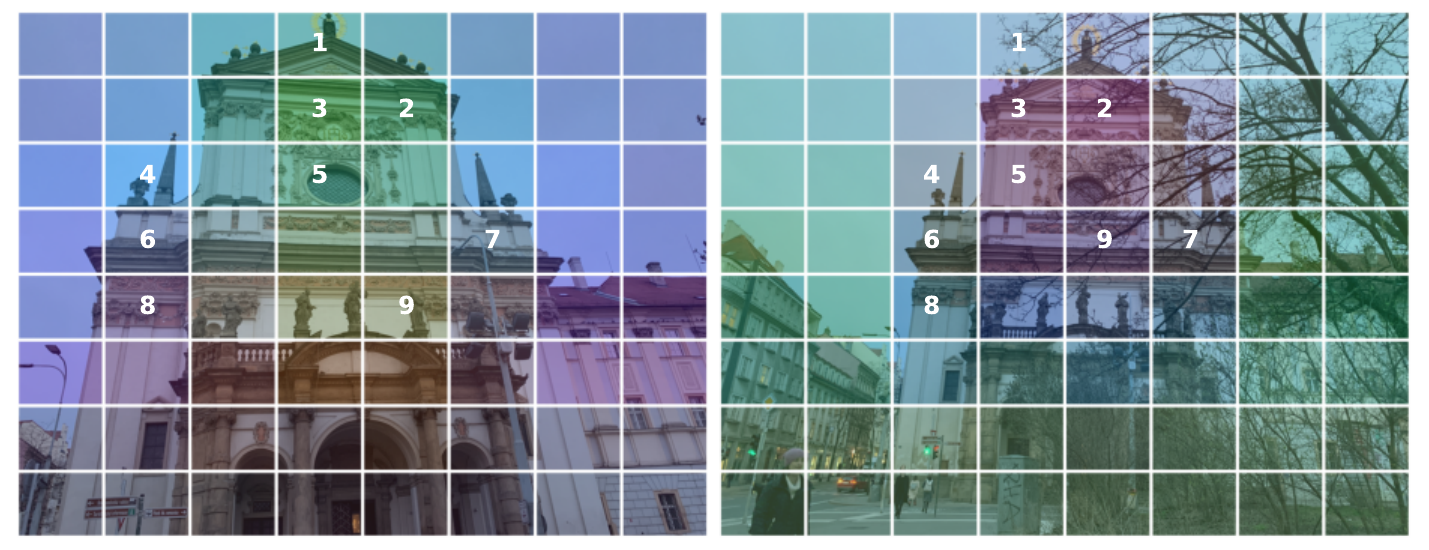}
         \vspace{-0.5em}
    \caption{
    A patch matching example with a patch number of $8^2$. 
    The 256 patches from DINOv2~\cite{oquab2023dinov2} are average pooled to 64. The numbers inside the patches indicate which ones are matched. The color overlay is calculated by PCA on patch embeddings. }
         \vspace{-0.5em}
    \label{fig:examples_matching}
\end{figure}
\begin{equation}
\begin{split}
    o'(p, q) = \sum_{X \in \mathcal{X}} \llbracket 
    \text{inside}(\mathbf K_i (\mathbf R_i \mathbf X + \mathbf t_i), p) \rrbracket * \\
   \llbracket  \text{inside}(\mathbf K_j (\mathbf R_j \mathbf X + \mathbf t_j), q) \rrbracket,
\end{split}
\label{eq:overlap2}
\end{equation}

\noindent
where $\mathcal{X} \subset \mathbb{R}^3$ comprises 3D points from a 3D reconstruction of the environment, formula $\mathbf K (\mathbf R \mathbf X + \mathbf t)$ projects a 3D point to the image, function $\text{inside}(\cdot)$ checks whether a 2D point falls into a 2D patch, 
$\llbracket \cdot \rrbracket$ is the Iverson bracket which equals one if the condition inside holds and zero otherwise.

Eq.~\ref{eq:overlap2} defines the patch overlap as the number of 3D points commonly visible on both patches. 
Then the overlap of two entire images $I_i$ and $I_j$ is defined as the sum of overlap scores across matching patches as follows:
\begin{equation}
    O(I_i, I_j) = \sum_{p \in \mathcal{P}_i} \sum_{q \in \mathcal{P}_j} o'(p, q) = \sum_{(p, q) \in \mathcal{C}} o'(p, q),
\end{equation}
where set $\mathcal{C} = \{ (p, q) \; | \; p \in \mathcal{P}_i, q = \arg_w \max o'(p, w) \in \mathcal{P}_j, w \in \mathcal{P}_j \}$ is a set of patch-to-patch correspondences defined by assigning the most similar patch from the database image to each patch in the query.

However, this formulation still does not allow for retrieval, as the camera pose corresponding to the query image is unknown in practice.
Thus, we further approximate the image overlap as the similarities between the patch embeddings in a robust way as follows:
\begin{equation}
    \hat{O}(I_i, I_j) = \sum_{(p, q) \in \mathcal{C}} \rho (sim(f(p), f(q))),
\end{equation}
where 
$f(\cdot)$ is an encoder with a single patch as input, usually using deep networks or hand-crafted methods,  
and $\rho : \mathbb{R} \to \mathbb{R}$ is a robust function, \eg, $\rho$ can be defined as a voting mechanism as $\rho(x) = \llbracket x > \epsilon \rrbracket$, where $\epsilon$ is a manually set threshold with $x$ as the patch similarity $sim(f(p), f(q))$. 

Given such a scheme to quantify image overlap via the similarity of a set of patch correspondences, 
the most similar image $I^*$ that we retrieve for query $I_i$ from a set of database images $\{ I_j \},  j \in 1, \dots, N$ is the one with the highest overlap, calculated as follows:
\begin{equation}
    I^* = \arg_{j \in 1, \dots, N}  \max \hat{O}(I_i, I_j).
\end{equation}
In the scenario where multiple images need to be retrieved for a single query, we straightforwardly define a ranking by sorting images with high visual overlap.
%
Thus, image retrieval from collections becomes a patch-matching problem.
Example matches are visualized in Fig.~\ref{fig:examples_matching}.
%
Next, we will talk about how to learn the encoder $f(\cdot)$ in VOP focusing on accurate overlap prediction.

\subsection{Learning Patch Matching via Patch Similarities}

In order to find the similarity of image patches $p$ and $q$ in images $I_i$ and $I_j$, respectively, we aim to learn an embedding $f(\cdot)$ that allows for casting the problem as a descriptor matching.
Thus, we aim to distill embeddings $e_p \in \mathbb{R}^d$ and $e_q \in \mathbb{R}^d$ such that $sim(p, q) (= \delta(e_p, e_q) =  \delta(f(p), f(q))$) is high if and only if $p$ and $q$ are overlapping and zero otherwise. 
The function $\delta(\cdot)$ measures the similarity in the embedding space, \eg, as the cosine similarity. 
\begin{figure*}[ht]
    \centering
    \vspace{-0.5em}
    \includegraphics[width=0.92\textwidth, trim = 0mm 0mm 0mm 0mm, clip]{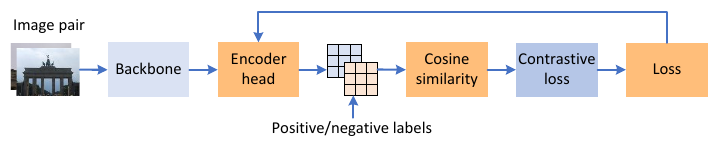}
    \vspace{-0.5em}
    \caption{
    The training pipeline of the proposed Visual Overlap Prediction (\textbf{VOP}) model, including a frozen DINOv2 backbone~\cite{oquab2023dinov2} breaking down input images into rectangular patches, a trainable encoder head, and contrastive loss using patch-to-patch overlap supervision.
    }
    \vspace{-0.5em}
    \label{fig:pipeline}
\end{figure*}
\begin{figure*}[ht]
    \centering
    \includegraphics[width=0.92\textwidth, trim = 0mm 0mm 0mm 0mm, clip]{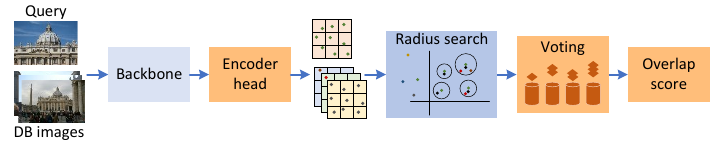}
    \vspace{-0.5em}
    \caption{
    The proposed \textbf{VOP} at inference.
    Given an input query image and a reference image collection,
    the frozen backbone~\cite{oquab2023dinov2} extracts patch-level features, 
    which are then fed into our trained encoder to obtain the final embeddings.
    For each patch in the query image, the radius neighbor search is performed in the embedding space.
    The final overlap scores are determined by robust voting on the formed ``query''-to-``database'' patch neighbors.
    In practice, the map embeddings are offline pre-calculated and saved.}
    \label{fig:inference}
    \vspace{-0.5em}
\end{figure*}

The \textbf{pipeline} for learning patch-level embeddings is visualized in Fig.~\ref{fig:pipeline}.
Given a pair of images $I_i$ and $I_j$, the goal is to learn embeddings $e_p$ and $e_q$ as defined previously.  
First, we obtain patch features by inputting the images into a frozen DINOv2~\cite{oquab2023dinov2} backbone.
%
These features are then fed into an encoder head, comprising a linear projection layer plus a few fully connected layers with GELU~\cite{hendrycks2016gaussian} activation and a dropout layer to encode the centers of the patches.
Given the corresponding patch pair $(p, q)$ as a positive sample, we train the encoder to predict similar embeddings for $p$ and $q$.
To do so, we employ contrastive learning on the similarities over the learned embeddings.

\vspace{1mm}
\noindent
\textbf{Contrastive Learning.}
We use contrastive learning to learn a joint embedding space for the image patches. 
To do so, we form patch pairs $(p, q)$. 
Real-world scenes are rarely static, \eg, objects move or undergo non-rigid deformations and illumination changes.
To ensure that the learned embedding is robust to such temporal changes,
we use image augmentation techniques as implemented in GlueFactory~\cite{lindenberger2023lightglue}, such as adding random brightness, blur, flip, and noise.

For each query patch $p$, we prepared a set of candidate patches $\{q, q_1, q_2, ..., q_n \}$ for training, where $\{ q_1, q_2, ..., q_n \}$ act as $n$ negative samples, depicting patches that do not have a common field-of-view with the query patch. 
In our training pipeline, 
half of the samples have no overlap (negative samples), \ie, they depict different scenes. 
The other half are from the same scene as the query. 
In the positive image samples, there are 30-90\% negative patches, which are the ones not matchable with the query patch. 
We consider the loss in a patch-level manner.
%
We train our model by optimizing the contrastive loss on the similarities of the embeddings of patch-to-patch matches:
%
%
%
  
%
\begin{equation}
\begin{split}
    \mathcal{L}_\text{contrastive} = \frac{1}{n^2}\sum_i^n \sum_j^n  \biggl( l_\text{GT} * (\sigma - \delta(e_p, e_q))^2 \\
    -  (1-l_\text{GT}) * \delta(e_p, e_q)^2 \biggr) ,
    \end{split}
    \label{eq:contrastive}
\end{equation}
where $l_\text{GT}$ is the ground-truth labels for each pair of patches (one from the query image and the other from the database). It is positive if their similarity is larger than a threshold $\llbracket sim(p_{i}, q_{j}) > \epsilon\rrbracket$ and vice visa.
Parameter $\delta(\cdot)$ is the similarity measurement of the $d$-dimensional descriptors, \ie, cosine similarity with $\sigma = 1$ as the margin.
It penalizes the cases of high similarities on the patches but non-overlapping in the ground truth, \ie, no visible 3D points. 
We resize the input images to a fixed size. 
Therefore, the overall loss is averaged over $n^2$ tentative matching patches on one image pair. 
We also experimented with incorporating attentions into the procedure.  
However, we noticed that the attention layers only increase the size of the network while leading to worse generalization performance. 
Thus, we keep only the contrastive loss for training.

%

To provide means for quick prefiltering, we add a classification ([CLS]) token to the employed Transformer network that is trained to distill global image embeddings based on the image overlap scores, shown in the ablations of the Supp. Mat.
These tokens are used to efficiently select a subset of potentially overlapping images given the query. 

\vspace{1mm}
\noindent
\textbf{Supervision.}
Now, we introduce the visual supervision built for training. 
The ground-truth overlapping patches of the training image pairs are achieved 
following LightGlue~\cite{lindenberger2023lightglue} employing pixel-wise depth and relative pose. 
As shown in Eq.~\ref{eq:overlap2}, the patch overlap is measured by counting the co-visible 3D points in the reconstruction.
The ground-truth matching local patches are found in two steps:
First, a set of dense pixel coordinates $c_i$ are detected in the first image $I_i$, and transformed to 3D points $X_i$.
The obtained 3D point cloud is filtered by non-zero depths and transformed to the reference camera, \ie, projecting the keypoints from camera $i$ to $j$ by the given transformation matrix, $kp_{i 2 j}$.
Given camera intrinsics, we project the 3D points $X_i$ to the camera planes and check for visibility.
We make sure that only cycle-consistent points are kept by using the projected point, calculating its 3D coordinates, and back-projecting to the original image for a visibility check again.
The ground-truth matches are supposed to contain the points with a consistent depth~\cite{lindenberger2023lightglue}. 
This procedure provides the initial set of positive and negative pairs of matching image patches. 

Next, given the visible pixel matches, we check which patches they belong to and define the matchable patches by having at least one visible pixel correspondence.  
We assume one-to-one matching relations, \ie, the patch with the most patch matches in the image pair will be chosen if there are multiple. 
Note that the visual supervision is built mutually from the query image to the reference and vice versa. 

\subsection{Image Retrieval by Overlap Scores}

Given a query image $I_i$ and a set of database images ${I_j}_{j = 1, ..., N}$, our goal is to efficiently determine the top-$k$ images for $I_i$ maximizing overlap.
To do so, we pre-generate the patch embeddings of all images in the database during the offline mapping phase of the method. 

The retrieval pipeline is visualized in Fig.~\ref{fig:inference}. 
In the online phase, we obtain the patch and, optionally, the global embeddings of the query image.
Next, for each query image patch, we perform a radius search in the database of patch embeddings to determine all potentially overlapping candidates.
We use the rounded median similarity over 100 random samples as the radius threshold.
Note that k-nearest-neighbors algorithm is insufficient in our case as we would like to find all potentially overlapping patches to determine image overlaps.
Such a radius search can be efficiently performed by standard bounding structures, \eg, kd-tree, and other implementation tricks.
For efficiency, we also use the global descriptor to prefilter the images to a shortlist. 

To obtain the top-$k$ images, we perform voting such that the overlap calculation becomes:
\begin{equation}
    \hat{O}(I_i, I_j) = \sum_{(p, q) \in \mathcal{C}} \llbracket sim(p, q) > \epsilon \rrbracket,
\end{equation}
quantifying the image overlap as the count of overlapping patches inside the images, or the sum of the similarities of the descriptors as
\begin{equation}
\hat{O}(I_i, I_j) = \sum_{(p, q) \in \mathcal{C}} \max(\delta(e_p, e_q) - \epsilon, 0).
\end{equation}

\vspace{1mm}
\noindent
\textbf{Weighted Voting.}
%
To highlight the distinguishable query patches, we propose to weight the query patches from each image using a bag of visual words approach~\cite{csurka2004visual}. Specifically, we adapt TF-IDF~\cite{ramos2003using} weights to all the query patches, assigning lower weights to patches that frequently appear in queries but are rare in the matched database patches.
As illustrated below, it is preferable to have more neighbor patches if fewer images share the same neighborhood with the query patch. For each query patch, the weight $t_i$ is calculated as follows:
\begin{equation}
    t_i = \frac{n_{id}}{n_d} \log \frac{N}{n_i},
\end{equation}
%
where $n_{id}$ is accumulated over all patch neighbors in a one-to-many manner, \ie, not forced to find only one neighbor in each database image.
Patch neighbors are defined as the cosine similarities of the patch descriptors within a specified radius.
Parameter $n_d$ is the total number of candidate patches in all DB images, \ie, 256 * $N$ in our case, with a patch size of $14^2$ on the resized image size of $224^2$.
Parameter $N$ represents the number of DB images, and $n_i$ denotes the number of database images that contain at least one neighbor of the query patch. 
Patches that frequently appear are deemed less important and thus are down-weighted.
Weight $t_i$ for each query patch is incorporated into the overlap score computation by $t_i *  \hat{O}(I_i, I_j)$. 
Note that the map weights are precomputed during the mapping phase.

\section{Experimental Results}

Our experiments include estimating two-view epipolar geometry and localizing query images from retrieved images. The VOP model is trained on 153 scenes from the MegaDepth dataset~\cite{MDLi18}, which includes Internet-sourced images and depths reconstructed via Structure-from-Motion (SfM) and multi-view stereo techniques. 
Beyond two MegaDepth testing scenes, we evaluate the generalization ability of the proposed VOP on 11 scenes from PhotoTourism~\cite{snavely2006photo}, 13 scenes from ETH3D~\cite{schops2017multi}, and the InLoc dataset~\cite{taira2018inloc}.
We benchmark the proposed method against several competitors, including CNN-based retrieval methods like NetVLAD~\cite{arandjelovic2016netvlad} and CosPlace~\cite{berton2022rethinking}, as well as the state-of-the-art (SOTA) retrieval method AnyLoc~\cite{keetha2023anyloc} and DINOv2-SALAD~\cite{Izquierdo_CVPR_2024_SALAD} utilizing the same DINOv2 backbone as ours.
Besides those \textit{global-only} methods, 
we compare {\bf{VOP}} with the SOTA reranking-based methods: the ViT-based $R^2$Former~\cite{Zhu_2023_R2Former} and Patch-NetVLAD~\cite{Hausler2021}. 
Also, we finetune CosPlace* on the same training data as ours to show the improved performance is not due to the different training data. 
In addition, we test the reranking methods~\cite{Hausler2021} \cite{Zhu_2023_R2Former} on the shortlists created by DINOv2 [CLS] tokens, same as VOP, marked with $\dagger$ in the tables.


\vspace{1mm}
\noindent
\textbf{Training Details.}
The models underwent training on the 153 MegaDepth scenes, employing contrastive loss to differentiate negative and positive patch pairs. Each scene contributes 150 positive and 150 negative image pairs to the training set, standardized to an image resolution of $224^2$ pixels. We use the pre-trained DINOv2 with a patch size of $14^2$ to extract $1024$-dimensional descriptors and reduce them to 256 dimensions by a fully-connected layer, \ie, the first layer of our training encoder head.
Negative images, characterized by zero overlaps (\eg, sourced from distinct scenes), contrast with positive samples randomly selected from the pairs with 10\% to 70\% overlapping patches. 
Notably, the contrastive loss is predicated on patch-level negative/positive distinctions, \ie, cosine similarities, rather than on entire images. 
Following LightGlue~\cite{lindenberger2023lightglue}, the patch-level supervision is generated from depth and visual features.
The training was conducted in 30 epochs with a batch size of 64, employing the validation losses as the primary criterion for best checkpoint selection with a learning rate of $1e-4$. 

\vspace{1mm}
\noindent
\textbf{Evaluation Metrics.}
%
In the evaluation, 
we use the proposed weighted voting procedure with a radius search thresholded at the rounded median similarity over 100 random image samples.
Note that the TF-IDF weights are used only when they are non-zero.
Then, we assess the relative poses between the query and the top-$k$ retrieved images using robust estimation techniques. 
This involves deploying SuperPoint~\cite{DeTone_2018_CVPR_Workshops} and the state-of-the-art LightGlue~\cite{lindenberger2023lightglue,sarlin2019coarse} for feature detection and matching, followed by OpenCV-RANSAC~\cite{RANSAC} running for 10K iterations to compute the relative poses. 
The performance is measured against ground truth (GT) poses by calculating the Area Under the Recall curve (AUC) thresholded at $10^\circ$~\cite{cne2018, brachmann2019neural}, along with the median pose error. 
The RANSAC threshold is automatically selected from a set of pixel values $\{0.5, 1, 2, 3\}$ following GlueFactory~\cite{lindenberger2023lightglue}.
Also, we report the average number of inliers across the retrieved image pairs found by RANSAC. 
Note that the number of inliers is not an indicator of the accuracy of the estimated pose, as discussed and verified in  MQNet~\cite{barath2022learning} and FSNet~\cite{barroso2023two}.  
In scenarios where $k>1$, for each query, we follow the practical approach and select only the image pair with the highest number of inliers. 
This comprehensive evaluation allows us to thoroughly assess the performance not just in retrieving relevant images but also in accurately estimating their relative poses.

Additionally, we run hloc~\cite{sarlin2019coarse} on the top-40 retrieved images and evaluate the absolute poses of the queries using the long-term visual localization benchmark~\cite{toft2020long}.
We compare the percentages of the correctly estimated translations and rotations with errors below 0.5$m$ and 5$^\circ$.

\begin{table}[t]
    \centering
    \resizebox{1\columnwidth}{!}{
    \begin{tabular}{r | c c c | c c c | c c c }
    \hline
       & \multicolumn{3}{c|}{AUC@10$^\circ$~$\uparrow$} & \multicolumn{3}{c|}{Med.\ pose error ($^\circ$)~$\downarrow$} & \multicolumn{3}{c}{\# inliers~$\uparrow$} \\
        Top $k$ $\rightarrow$ &   \phantom{---}1\phantom{---} & \phantom{---}5\phantom{---} &  \phantom{---}10\phantom{---} &  \phantom{---}1\phantom{---} &   \phantom{---}5\phantom{---} &  \phantom{---}10\phantom{---} &   \phantom{---}1\phantom{---} &   \phantom{---}5\phantom{---} &   \phantom{---}10\phantom{---}\\\hline
NetVLAD~\cite{arandjelovic2016netvlad}  &56.5 &64.2&63.6 &2.85 &2.36 &2.38 &234.0 &\bf{286.5} &294.0\\
CosPlace~\cite{berton2022rethinking} &54.5&63.5& 64.3 &3.14 &2.34 & 2.27 &\bf{239.5} &285.0 &\underline{\bf{294.5}}\\
CosPlace*~\cite{berton2022rethinking} &59.3&  65.8 &66.2&2.58 &2.65 &\bf{2.10} &164.0&162.0&255.5\\
DINOv2~\cite{oquab2023dinov2} &61.1&65.6&64.8 &2.32 &2.29 &2.31  &219.0 &270.0 &282.5\\
AnyLoc~\cite{keetha2023anyloc}&60.8& 65.3 &64.1&2.38 &2.26 &2.35 &220.5 &273.0 &287.5\\
 SALAD~\cite{Izquierdo_CVPR_2024_SALAD} & 53.9 &61.8 &62.6 &3.23 & 2.61 &2.54 & 224.5 &278.0 &285.0 \\
P-NetVLAD~\cite{Hausler2021} &59.7 & 64.9 & 64.5  & 2.67  & 2.22 & 2.27 &   212.5 & 264.0 & 283.5\\
$\dagger$P-NetVLAD~\cite{Hausler2021} &62.3 &64.5 &65.1 &2.56 &2.31 &2.21 &212.5 &262.0 &275.5\\
$R^2$Former~\cite{Zhu_2023_R2Former}& 57.9 & 65.4 & 65.7 & 3.03 & 2.24  & 2.21 & 108.0 &  185.5 & 214.0\\
$\dagger$$R^2$Former~\cite{Zhu_2023_R2Former}& \bf{64.2} & \underline{\bf{67.9}} & 65.9 & \bf{2.25} &2.05 & 2.27 &186.0 &247.0 &273.5\\
 \bf{VOP} & 61.8 &67.6 & \bf{66.7}& 2.36 &\underline{\bf{2.02}} &2.22 &177.5 & 246.5 & 251.0\\

\hline
    \end{tabular}}
         \vspace{-0.5em}
    \caption{
	Pose AUCs and median errors on MegaDepth~\cite{MDLi18} using top-$k$ images retrieved by different methods, with the best results in bold at each $k$ and the overall best underlined.  $^\dagger$ results are tested by reranking images prefiltered by the [CLS] tokens from DINOv2. * shows results of fine-tuned CosPlace on Megadepth.}
 \vspace{-0.5em}
    \label{tab:megadepth}
\end{table}

\subsection{Pose Estimation on MegaDepth}
Table~\ref{tab:megadepth} presents the results on the MegaDepth.
%
%
The proposed method secures comparable scores for the AUC@10$^\circ$ criterion and exhibits the lowest median pose errors among the SOTA baselines.
The underlined results mark the optimal configuration for relative pose accuracy, evidencing the superior performance of the proposed VOP approach when utilizing the top-5 images. 
This configuration outperforms the global-only retrieval baselines (the first seven rows), reinforcing the significance of selecting correct metrics and model configurations tailored to the specific demands of real-world downstream applications.
Our method outperforms the recent reranking approaches $R^2$Former~\cite{Zhu_2023_R2Former} and Patch-NetVLAD~\cite{Hausler2021} on MegaDepth when using their pretrained models. 
When using the same shortlist we use, their performance (with $\dagger$) is marginally better than ours. 
%
The fine-tuned CosPlace (CosPlace*) achieves better accuracy than the pretrained model, especially achieving good results on top-10 pairs. 
Next, we will show the generalization and robustness of the compared methods on unseen data. 

\begin{table}[t]
    \centering
    \resizebox{1\columnwidth}{!}{
    \begin{tabular}{r|c c c |c c c | c c c }
    \hline
      &\multicolumn{3}{c|}{AUC@10$^\circ$~$\uparrow$} & \multicolumn{3}{c|}{Med. pose error~($^\circ$)~$\downarrow$} & \multicolumn{3}{c}{\# inliers~$\uparrow$} \\
    Top $\rightarrow$ &    \phantom{---}1\phantom{---} & \phantom{---}5\phantom{---} &  \phantom{---}10\phantom{---} &  \phantom{---}1\phantom{---} &   \phantom{---}5\phantom{---} &  \phantom{---}10\phantom{---} &   \phantom{---}1\phantom{---} &   \phantom{---}5\phantom{---} &   \phantom{---}10\phantom{---}\\\hline
NetVLAD~\cite{arandjelovic2016netvlad} & 89.5 &90.4 &89.8 &0.79 &0.77 &0.82&378.4 &419.3&444.9\\
CosPlace~\cite{berton2022rethinking}& 88.3 &89.8&89.8 &0.82 &0.80 &0.80&352.3 &421.6 &434.9 \\
CosPlace*~\cite{berton2022rethinking} & 43.6&79.7&85.9 & 38.3 & 0.79& 0.76 & \phantom{1}62.6 & 279.0 & 360.5\\
DINOv2~\cite{oquab2023dinov2}  & 87.2 & 89.9 &89.6&0.82 &0.83 &0.86 &365.2 &415.0 &436.7 \\
AnyLoc~\cite{keetha2023anyloc}&  88.9&90.2&90.2&0.77& 0.83 &0.80&\bf{373.2} &\bf{425.1}&438.0\\
 SALAD~\cite{Izquierdo_CVPR_2024_SALAD} & \bf{89.7} &90.2 &90.0& \underline{\bf{0.68}} &0.76 &0.78 &385.9 &411.2 &442.0\\
 P-NetVLAD~\cite{Hausler2021} & 86.7 & 88.4 & 90.2 & 0.81 & 0.83 & 0.81 & 333.3 & 412.4 & \underline{\bf{466.8}}\\
  $^\dagger$P-NetVLAD~\cite{Hausler2021} & 80.5 & 88.1 &89.4 & 0.84  & 0.77 & 0.74 & 291.7  & 341.5 & 372.2\\
$R^2$Former\cite{Zhu_2023_R2Former} & 50.2   & 72.0& 84.4 & 34.0 & 1.58 & 0.84& 137.0 &318.3&401.5\\
$^\dagger$$R^2$Former\cite{Zhu_2023_R2Former} &51.1  &83.0 &87.4&13.4  &0.79 &0.86&114.7 &272.5 &314.7\\
 \bf{VOP} & \bf{89.7} &  \underline{\bf{90.5}} & \bf{90.4} &0.69 & \bf{0.70} & \bf{0.70} &341.1 & 399.5 &430.7 \\
\hline
    \end{tabular}}
     \vspace{-0.5em}
    \caption{Relative pose AUC@10$^\circ$ and median errors on the unseen ETH3D dataset~\cite{schops2017multi} on the top-$k$ retrieved images.
	}
 \vspace{-0.5em}
    \label{tab:eth3d}
\end{table}
\subsection{Generalization Experiments }
\label{sec:generalization}

We test the proposed method on multiple real-world scenes from the ETH3D~\cite{schops2017multi} and PhotoTourism~\cite{snavely2006photo} datasets for relative pose estimation, and InLoc~\cite{taira2018inloc} for localization.
We use the model trained on MegaDepth, \ie, no fine-tuning or retraining is done on the following datasets.

\vspace{1mm}
\noindent
\textbf{ETH3D}~\cite{schops2017multi} benchmark consists of images with various viewpoints from \textit{indoor} and \textit{outdoor} scenes, with ground-truth poses reconstructed by laser scanner. 
13 scenes are used for testing.
All images in the dataset are used as queries to select database images from the rest. 
Table~\ref{tab:eth3d} shows that {\bf{VOP}} is always among the top-performing methods. 
The overall best results in all accuracy metrics are always obtained by VOP by selecting top-5 images.
This highlights that VOP performs accurately even without retraining and, thus, is applicable to unseen scenes. 
The fine-tuned CosPlace does not generalize well to ETH3D, and yields degraded results, especially on the top-1.
The reranking methods, P-NetVLAD and $R^2$Former, significantly lag behind the performance of VOP, both in their original version and when combined with DINOv2 [CLS] tokens.  

\vspace{1mm}
\noindent
\textbf{PhotoTourism.}
Table~\ref{tab:photo} reports the relative pose estimation results on 11 testing scenes from the PhotoTurism~\cite{snavely2006photo} dataset. 
The proposed VOP achieves the lowest median pose error and the second-best AUC scores. 
Similar to Tables~\ref{tab:megadepth} and \ref{tab:eth3d}, Table~\ref{tab:photo} shows that the number of inliers is not an indicator of accuracy, as we result in fewer inliers while being more accurate than baselines. 

\begin{table}[t]
    \centering
    \resizebox{0.99\columnwidth}{!}{
    \begin{tabular}{r|c c c | c c c | c c c }
    \hline
            &\multicolumn{3}{c|}{AUC@10$^\circ$~$\uparrow$} & \multicolumn{3}{c|}{Med. pose error ($^\circ$)~$\downarrow$} & \multicolumn{3}{c}{\# inliers~$\uparrow$} \\
    Top $\rightarrow$ &    \phantom{---}1\phantom{---} & \phantom{---}5\phantom{---} &  \phantom{---}10\phantom{---} &  \phantom{---}1\phantom{---} &   \phantom{---}5\phantom{---} &  \phantom{---}10\phantom{---} &   \phantom{---}1\phantom{---} &   \phantom{---}5\phantom{---} &   \phantom{---}10\phantom{---}\\\hline
NetVLAD~\cite{arandjelovic2016netvlad} & 55.0  & 59.3  & 60.3 & 3.75 & 3.10 & 2.94 & 283.9 & 340.0 &348.1\\
CosPlace~\cite{berton2022rethinking}& 53.4 &60.8&61.2 &3.78 &2.91 &2.87&\bf{291.5} &\bf{347.0} &354.1 \\
CosPlace*~\cite{berton2022rethinking} &58.3&65.9&66.1&3.59 &2.51 & 2.43&145.9&247.9&281.8\\
DINOv2~\cite{oquab2023dinov2} &61.1 &63.1 & 63.0 & 3.17 & 2.82 & 2.75 & 266.0  & 338.6 & \underline{\bf{356.3}}\\
AnyLoc~\cite{keetha2023anyloc}&  56.3&60.3&60.6&3.57& 2.90 &2.81&265.1 &334.6&348.6\\
 SALAD~\cite{Izquierdo_CVPR_2024_SALAD}&53.5&60.1 &59.6 &4.07 &3.19&3.17& 270.7 &338.1&349.7\\
 P-NetVLAD~\cite{Hausler2021} &61.9 &62.4 &61.8 &2.76 &2.65 &2.76 & 261.9 &315.2 &339.4\\
$^\dagger$P-NetVLAD~\cite{Hausler2021}  &62.1 &63.6 &62.8 &2.69 &2.63 &2.71 &246.5 &322.8 & 338.5\\
$R^2$Former~\cite{Zhu_2023_R2Former}&60.6 & \bf{67.8}& \underline{\bf{67.9}}&4.49 & 2.36 & 2.35 & 180.0 & 241.9 & 277.6\\
$^\dagger$$R^2$Former~\cite{Zhu_2023_R2Former}&\bf{63.1} & 63.9 &63.1&2.73 &2.55 &2.66 &216.4 &303.5 &331.6\\

  \bf{VOP} &62.5& 66.8 & 65.2 &\bf{2.63}&  \underline{\bf{2.12}} & \bf{2.44} & 224.5 & 250.5 &283.2 \\
\hline
    \end{tabular}}
         \vspace{-0.5em}
    \caption{
    Relative pose AUC@10$^\circ$ and median errors on the unseen PhotoTourism dataset~\cite{snavely2006photo} on the top-$k$ retrieved images.
	}
    \label{tab:photo}
\end{table}
\begin{table}[t]
    \centering
    \resizebox{0.999\columnwidth}{!}{
    \begin{tabular}{r| c c c c c c c c c}
    \hline
    \tabincell{r}{Method} & \tabincell{c}{NetVLAD\\~\cite{arandjelovic2016netvlad}}& \tabincell{c}{CosPlace\\~\cite{berton2022rethinking}} & \tabincell{c}{CosPlace*\\~\cite{berton2022rethinking}}  & \tabincell{c}{DINOv2\\~\cite{oquab2023dinov2}} & \tabincell{c}{AnyLoc\\~\cite{keetha2023anyloc}} &\tabincell{c}{SALAD\\~\cite{Izquierdo_CVPR_2024_SALAD}} & \tabincell{c}{$\dagger$P-NetVLAD\\~\cite{Hausler2021}} &\tabincell{c}{$\dagger$$R^2$Former\\~\cite{Zhu_2023_R2Former}} &   \tabincell{c}{\bf{VOP}} \\
    \hline
    DUC1 & 65.7  & 69.2 & 41.4 & 63.6 & \bf{74.7}& 71.2&60.1 &  47.0  & 72.2 \\
    DUC2 & 71.0 & 74.8 & 29.0 & 71.0 & 75.6 &\bf{78.6} &55.0  & 66.4 & 77.1 \\
    \hline
    \end{tabular}}
         \vspace{-0.5em}
    \caption{Indoor localization recalls at thresholds (5$^\circ$, 0.5$m$) on top-40 retrieved images from the InLoc~\cite{taira2018inloc} dataset.}
         \vspace{-0.5em}
    \label{tab: tuned}
\end{table}
\begin{table}[t]
    \centering
    \resizebox{1\columnwidth}{!}{
    \begin{tabular}{r| c c c c c c  c c c}
    \hline
    \tabincell{c}{Method} & \tabincell{c}{NetVLAD\\~\cite{arandjelovic2016netvlad}}& \tabincell{c}{CosPlace\\~\cite{berton2022rethinking}} & \tabincell{c}{CosPlace*\\~\cite{berton2022rethinking}}  & \tabincell{c}{DINOv2\\~\cite{oquab2023dinov2}} & \tabincell{c}{AnyLoc\\~\cite{keetha2023anyloc}} &\tabincell{c}{SALAD\\~\cite{Izquierdo_CVPR_2024_SALAD}}  &\tabincell{c}{$\dagger$P-NetVLAD\\~\cite{Hausler2021}} &  \tabincell{c}{$\dagger R^2$Former\\~\cite{Zhu_2023_R2Former}} &\tabincell{c}{\bf{VOP}} \\
    \hline
        Avg. acc.(\%)~$\uparrow$&70.1 &71.6&56.4&70.6&\underline{73.2}&72.3& 66.3&65.6 &\bf{74.8}\\
Avg. err.($^\circ$)~$\downarrow$ &\phantom{1}2.1&  \phantom{1}2.0 &\phantom{1}2.0&\phantom{1}2.0&\phantom{1}2.0&\phantom{1}2.2&\phantom{1}1.9&\phantom{1}\underline{1.8}&\phantom{1}\bf{1.6}\\
    \hline
    \end{tabular}}
         \vspace{-0.5em}
     \caption{
     Average accuracy (\%) and median pose error ($^\circ$) on the retrieved top-5 images on the Megadepth~\cite{MDLi18}, ETH3D~\cite{schops2017multi}, PhotoTourism~\cite{snavely2006photo}, and InLoc~\cite{taira2018inloc} datasets.
     InLoc is excluded from the median error as it is not available on the evaluation website.}
     \label{tab: avg}
     \vspace{-0.5em}
\end{table}

\begin{figure*}[ht]
    \centering
    \vspace{-0.5em}
    \includegraphics[width=0.495\textwidth, trim = 0mm 0mm 0mm 0mm, clip]{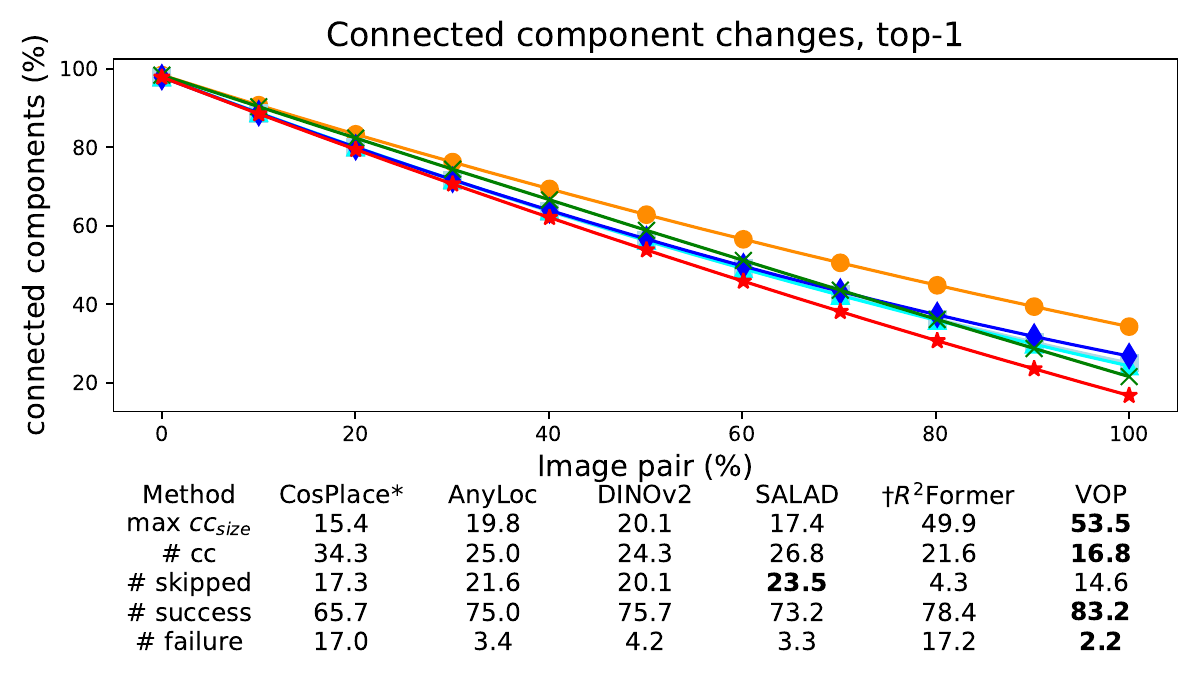}
    \includegraphics[width=0.495\textwidth, trim = 0mm 0mm 0mm 0mm, clip]{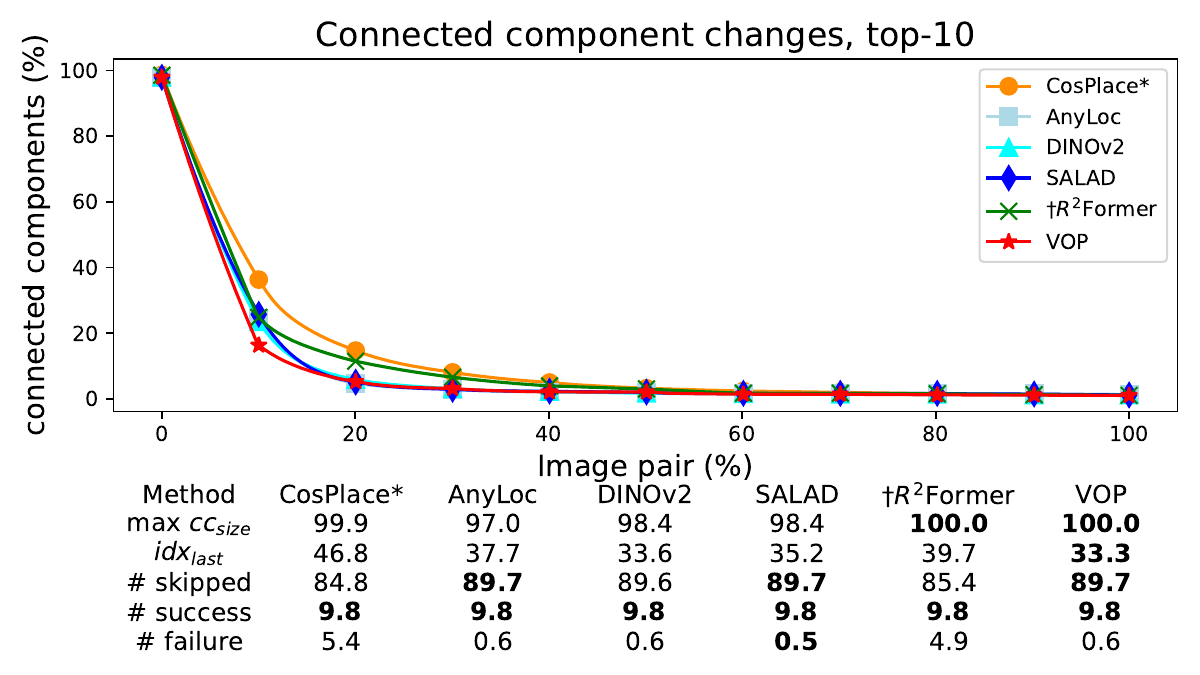}
    \vspace{-0.5em}
    \caption{
    The number of connected components (vertical axis, cc) is plotted against the index of the image pair (horizontal) on which RANSAC-based relative pose estimation runs. The \textit{left} plot shows results for the top-1 database (DB) images paired with $0.4K$ query images, where the number of pairs equals the number of queries. The \textit{right} plot shows results for the top-10 DB images with a termination criterion applied when all images are in a single component. 
    Row ``max $cc_{size}$'' is the number of elements in the largest cc. 
    Row ``\# cc'' is the final number of connected components, while ``$\text{idx}_{\text{last}}$'' shows the index when the termination criterion was triggered in the \textit{right} plot. 
Row ``\# skipped'', ``\# success'', and ``\# failure'' show the numbers of skipped, succeeded, and failed RANSACs. All are in percentages. }
    \vspace{-0.75em}
    \label{fig:graph}
\end{figure*}



\vspace{1mm}
\noindent
\textbf{InLoc.}
Additionally, we ran all methods on the InLoc dataset~\cite{taira2018inloc} (without fine-tuning) using hloc~\cite{sarlin2019coarse} for localization to show their generalization capabilities in the case of a large domain gap (in/outdoor). 
Following hloc~\cite{sarlin2019coarse}, we retrieve top-40 database images using global retrieval methods and the reranking ones (including VOP) with a top-100 shortlist created by [CLS] tokens.
Table~\ref{tab: tuned} shows the proposed {\bf{VOP}} achieves the second highest accuracy, showcasing its generalization abilities. 
Note that we only show results for P-NetVLAD and $R^2$Former with DINOv2 [CLS] tokens as their original version fails on this dataset.

\vspace{1mm}
\noindent
\textbf{In summary},
we show the average accuracy and median pose errors averaged over all tested datasets in Table~\ref{tab: avg}. 
Note that InLoc is excluded from the median pose error, as the official website~\cite{toft2020long} does not report it.
Overall, the proposed VOP leads to the highest accuracy and lowest pose errors. 
Fine-tuning CosPlace on Megadepth reduces its accuracy on average, demonstrating that our improved performance is not simply due to the different training data.

%

    

\begin{figure}[ht]
    \centering
        \vspace{-0.5em}
\includegraphics[width=0.495\textwidth]{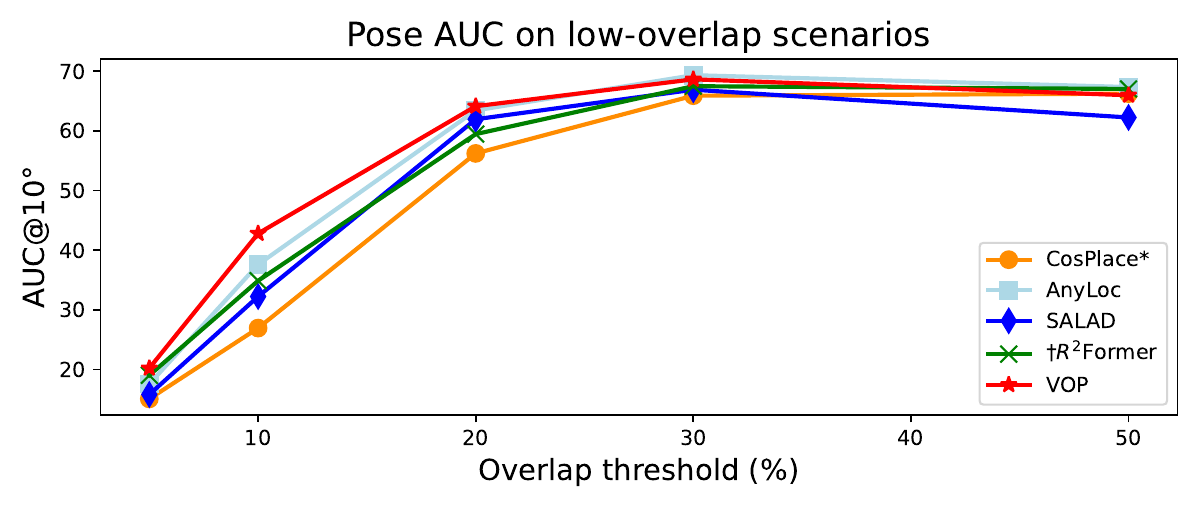}
\vspace{-1em}
    \caption{AUC@10$^\circ$ scores on the queries retrieving top-5 images from low-overlapping (\textit{horizontal axis}) database.}
    \label{fig:filter}
    \vspace{-0.75em}
\end{figure}
\subsection{Discussions}

\vspace{1mm}
\noindent
\textbf{Pose Graph Construction.}
We run a pose graph construction experiment using the retrieved image pairs in the following way.
All images are initialized to be connected components, each with a single element in it. 
We iterate through the queries and, for each, we add the query-to-database edge with the highest similarity/overlap to the reconstruction if and only if it connects two disconnected components by resulting in more than 10 inliers after running RANSAC. 
Thus, in each step, the number of connected components (cc) can only decrease by one or remain unchanged. Progressively, after all queries have been visited, we start again and process the edge with the second-highest similarity for each query. 
In the \textit{left (top-1)} plot of Fig.~\ref{fig:graph}, this is repeated until all edges have been visited; while termination criterion is used in the \textit{right (top-10)}, \ie, when all the images connected.
Note that we repeat such a process for 1K times with shuffling the queries and plot their average. 

The experiments were done on three test benchmarks: MegaDepth, PhotoTourism, and ETH3D. We show the averages by normalizing both horizontal and vertical axes. In the figures, the horizontal axis denotes the percentage of the currently processed image pairs (left - nothing processed so far, right - all pairs visited). The vertical axis is the percentage of connected components, \ie, the current number of cc divided by the initial.    
As shown in Fig.~\ref{fig:graph}, our method results in the fewest connected components with the largest max cc size at top-1 case. 
In the right plot, $R^2$Former and VOP are the only two methods that successfully connect all images in the pool, where VOP runs faster (more \# skipped RANSACs). 
VOP results in the fewest RANSAC failures at top-1 case, and second-best at top-10 with 0.1\% difference.

\vspace{1mm}
\noindent
\textbf{Difficult Scenario Experiment.}
We conducted an experiment on MegaDepth to retrieve the top-5 images in the low-overlap scenario. 
For this test, for each query image, we prefiltered the database by keeping only those images that have a lower overlap with the query than a threshold. 
Then, we ran image retrieval on this prefiltered database. 
%
As shown in Fig.~\ref{fig:filter}, VOP demonstrates clear improvements compared with other methods in this challenging scenario.

See the ablations of data augmentation, dropout layer and radius threshold sensitivity in the Supp. Mat. It also includes qualitative results and more discussions about storage and speed, patching sizes, supervision options, etc.

\section{Conclusion}

We introduce a novel Visual Place Recognition method, VOP, focusing on patch-level visual overlap prediction to address challenging viewpoint changes and occlusion. 
The proposed VOP predicts overlapping image patches by learning a patch-level descriptor and employing a robust voting mechanism and radius search in the descriptor space. 
Through extensive testing on real-world indoor/outdoor datasets and low-overlap scenarios, we demonstrate the effectiveness of VOP. 
On average, it achieves the highest accuracy, improving upon the state-of-the-art AnyLoc by two AUC points and reducing its median pose error by 20\%. 
Besides, VOP accelerates the pose graph construction process, tailored for geometric problems.
We believe this research marks a significant step forward in VPR problems by patch-level understanding, and it tailors such methods for 3D reconstruction from unstructured image collections.

\vspace{1mm}
\noindent
\textbf{Acknowledgements.}
D. Barath was funded by the ETH Zurich Career Seed Award. J. Matas was supported by the Technology Agency of the Czech Republic, project No. SS05010008 and project No. SS73020004.
T. Wei was funded by the Czech Science Foundation grant GA24-12697S and Research Center for Informatics (project CZ.02.1.01/0.0/0.0/16\_019/0000765 funded by OP VVV). 

{\small
\bibliographystyle{ieee_fullname}
\bibliography{egbib}
}

\section*{Appendix}

{\appendix 
\section{Ablations}
\vspace{1mm}

In this section, we will analyze certain components and parameters of the proposed method.

\vspace{1mm}
\noindent
\textbf{Radius Search Threshold.}
In the main paper, similar patches are selected by running radius search in the embedding space.
We compute the median similarity as the radius search threshold over 100 random samples from the query and database images.
Here, to understand how sensitive VOP is to the setting of this threshold, we show tuning results on the validation set of MegaDepth with different thresholds (horizontal axis) in Fig.~\ref{fig:threshold}.
AUC scores and median errors are shown on the top-10  retrieved image pairs.

\vspace{1mm}
\noindent
\textbf{Dropout Layer.}
Table~\ref{tab:ablation} shows the relative pose estimation performance on the top-$k$ images retrieved by the model with or w/o global prefilter ([CLS] tokens), and dropout layer.
Testing sets of MegaDepth are used, as in the main paper.
The [CLS] prefiltering improves the AUC scores. However, it increases the median pose errors marginally at the same time. 
As shown in the fifth row, data augmentation is essential in robustly learning the embeddings. 
Also, the last two rows in Table~\ref{tab:ablation} show that the dropout layer improves the performance on MegaDepth.
As shown in the main paper, VOP generalizes well for pose estimation on other data and indoor localization.

\begin{figure}[ht]
    \centering
    \includegraphics[width=0.85\linewidth, trim=0mm 0mm 0mm 0mm]{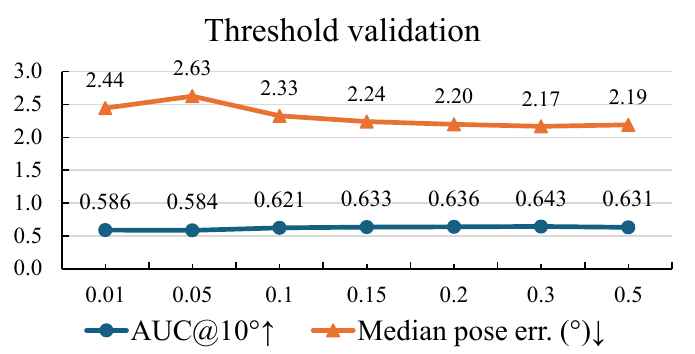}
    \caption{Ablations on the threshold used in radius search. AUC@10$^\circ$ and median pose errors on the validation scene shown.}
    \label{fig:threshold}
\end{figure}

\begin{table}[ht]
    \centering
    \resizebox{0.99\columnwidth}{!}{
    \begin{tabular}{ c c  c|c   c  c  }
    \hline
            Prefilter& Augment &Dropout&AUC@10$^\circ$~$\uparrow$ & Med. pose error ($^\circ$)~$\downarrow$ &  inliers~$\uparrow$ \\\hline
$\checkmark$ &$\times$ &$\times$ &65.1 &2.19 &\bf{272.5}\\
 $\times$ &$\checkmark$&$\times$ &66.3 &2.18 &222.0  \\
$\checkmark$ &$\checkmark$&$\times$ &66.7  &2.18 &263.0\\
$\checkmark$ &$\checkmark$&$\checkmark$ 
&\bf{67.6}
&\bf{2.02}
& 246.5 \\
\hline
    \end{tabular}}
         \vspace{-0.5em}
    \caption{
        Relative pose estimation on the MegaDepth dataset~\cite{MDLi18} on the top 5 retrieved images using different configurations, with the best results in bold.
        Prefilter indicates if the [CLS] token was employed to shortlist the potential candidates before overlap prediction.
        Augment refers to whether data augmentation was used.
	}
      \vspace{-0.5em}
    \label{tab:ablation}
\end{table}


\begin{figure*}[ht]
    \centering
    \begin{minipage}{\linewidth}
\includegraphics[width=0.99\linewidth, trim=2mm 3mm 3mm 2mm, clip]{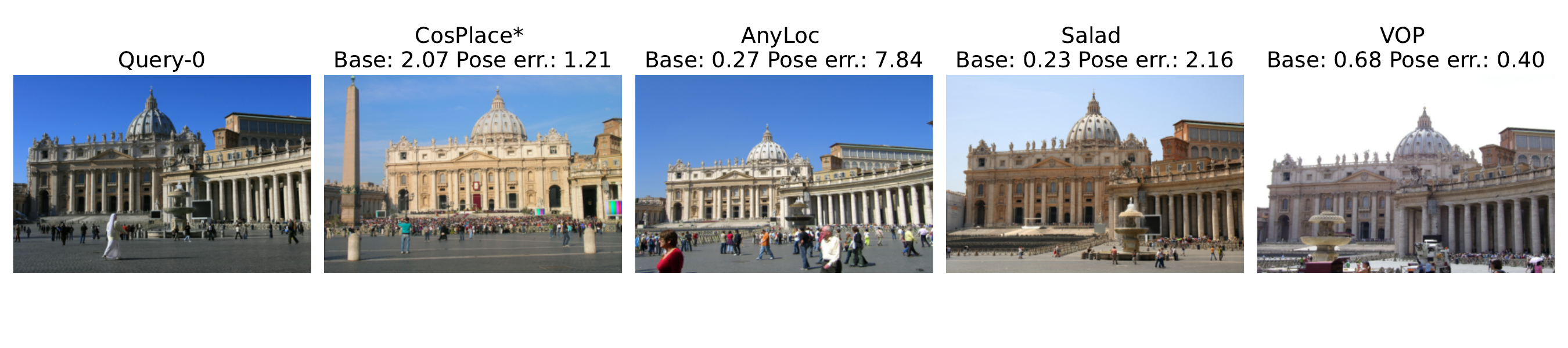}
\vspace{-2.5em}
\end{minipage}
\begin{minipage}{\linewidth}
\includegraphics[width=0.99\linewidth, trim=2mm 3mm 3mm 2mm, clip]{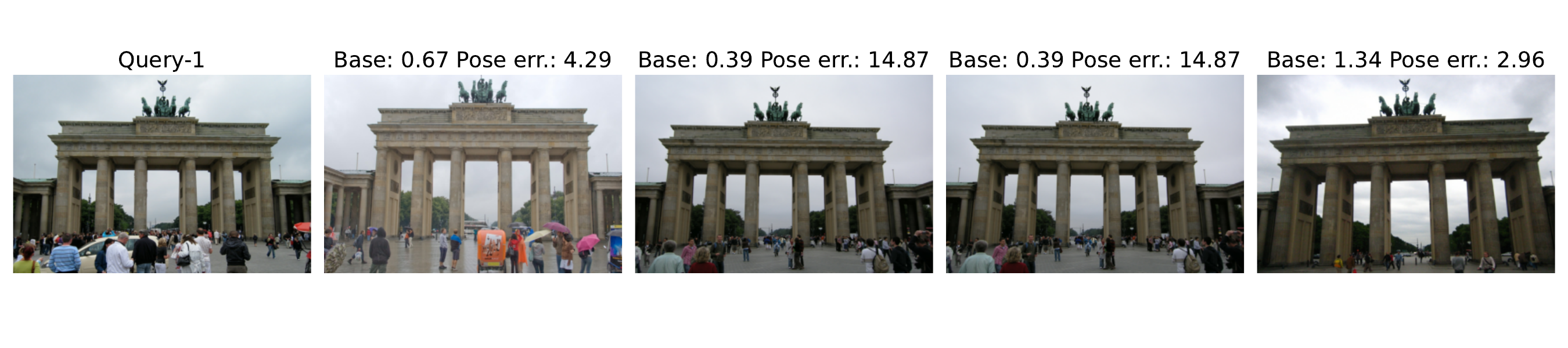}
\vspace{-2.5em}
\end{minipage}
\begin{minipage}{\linewidth}
\includegraphics[width=0.99\linewidth, trim=2mm 3mm 3mm 2mm, clip]{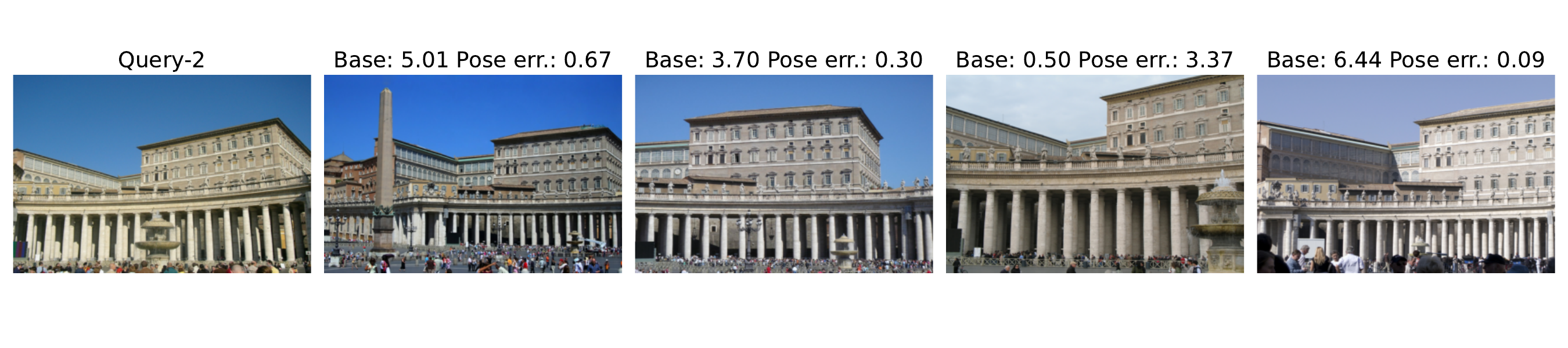}
\end{minipage}
\vspace{-2.5em}
    \caption{Baselines and pose errors between the retrieved images and queries using different methods shown.}
    \label{fig:baseline}
\end{figure*}

\begin{figure*}[t]
    \centering
    \includegraphics[width=1.95\columnwidth, trim=2mm 2mm 2mm 2mm ]{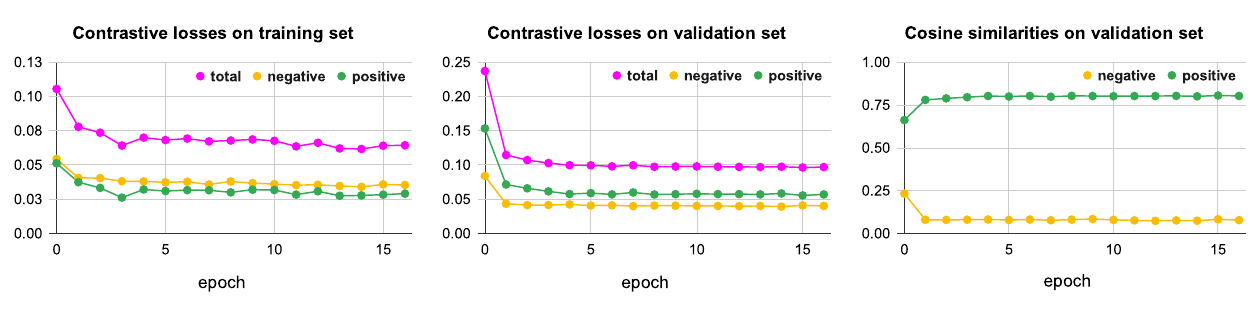}
    \vspace{-0.5em}
    \caption{Contrastive losses on the training set (\textit{left}) and validation (\textit{middle}) over different epochs shown for negative or positive patch pairs. 
    The \textit{right} plot shows the average cosine similarities over different patch samples.}
    \label{fig:loss}
\end{figure*}

\begin{figure*}[ht]
    \centering
    \vspace{-0.5em}
\includegraphics[width=0.495\textwidth, trim = 0mm 0mm 0mm 0mm, clip]{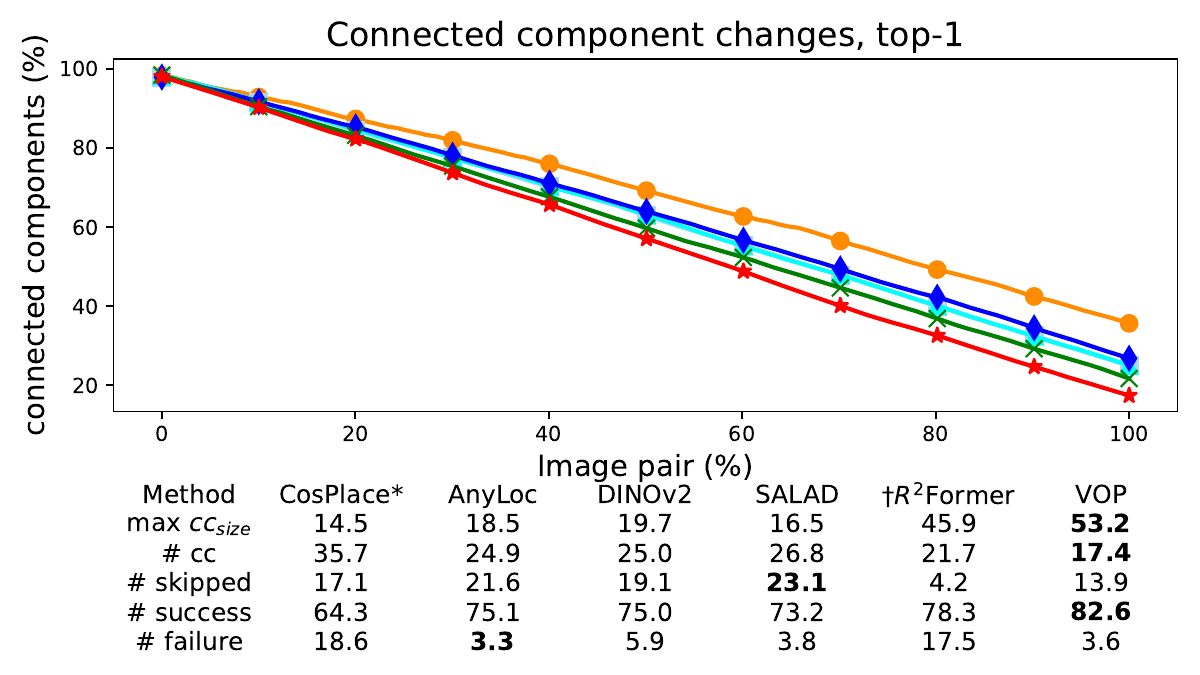}
    \includegraphics[width=0.495\textwidth, trim = 0mm 0mm 0mm 0mm, clip]{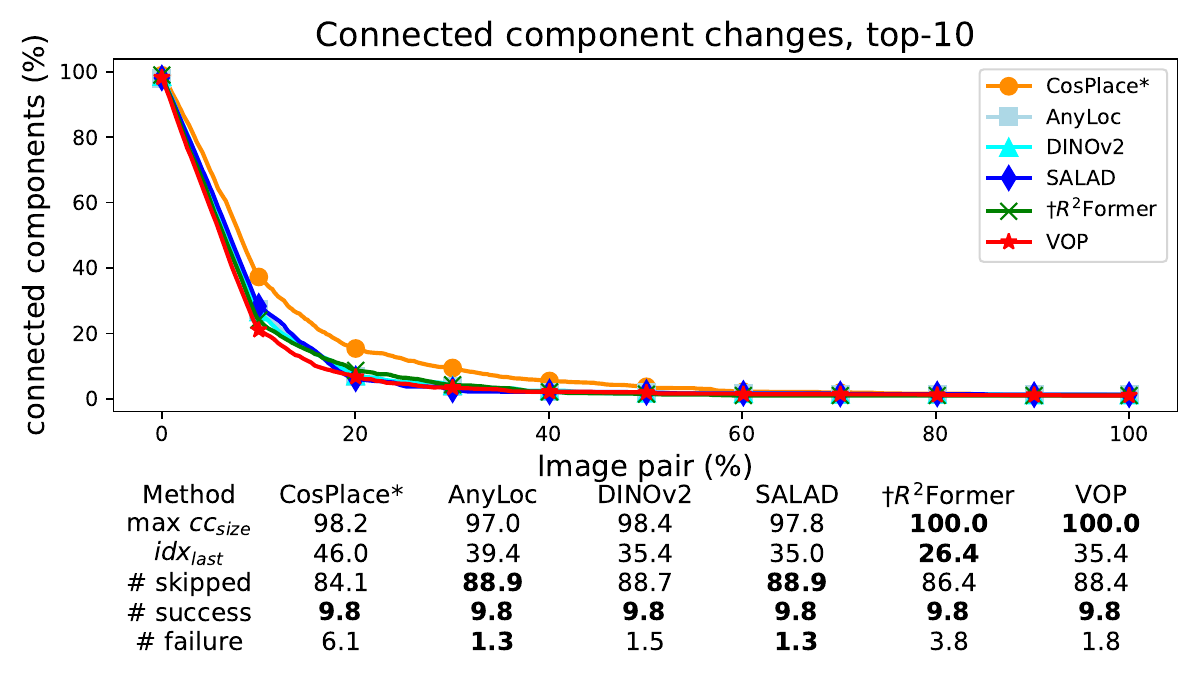}
    \vspace{-0.5em}
    \caption{
   The number of connected components (vertical axis, cc) is plotted against the index of the image pair (horizontal) on which RANSAC-based relative pose estimation runs. The \textit{left} plot shows results for the top-1 database (DB) images paired with $0.4K$ query images, where the number of pairs equals the number of queries. The \textit{right} plot shows results for the top-10 DB images with a termination criterion applied when all images are in a single component. 
    Row ``max $cc_{size}$'' is the number of elements in the largest cc. 
    Row ``\# cc'' is the final number of connected components, while ``$\text{idx}_{\text{last}}$'' shows the index when the termination criterion was triggered in the \textit{right} plot. 
Row ``\# skipped'', ``\# success'', and ``\# failure'' show the numbers of skipped, succeeded, and failed RANSACs. All are in percentages.
   }
    \label{fig:graph}
\end{figure*}


\section{Qualitative Results}
Most VPR methods prioritize retrieving similar images, typically resulting in short baselines that are not suitable for reconstruction. 
These goals conflict: the most similar images often produce short baselines, making pose estimation unstable. 
We aim to move beyond traditional similarity metrics and design retrieval methods tailored for geometric challenges, such as selecting images suitable for pose estimation. 
We visualize three query examples in Fig.~\ref{fig:baseline} with their top-1 retrieved images using different methods. 
VOP results in low pose errors as we find images with reasonable baselines for stable pose estimation.

\section{Discussions}

\noindent
\textbf{Training.}
To better understand the training process, Fig.~\ref{fig:loss} illustrates the training and validation losses on patch-level contrastive loss and the average similarity changing among different epochs on MegaDepth. 
It shows the contrastive loss helps to learn the embeddings of negative patches less similar and closer to positive ones, and it converges fast. 
The similarities shown are averaged over all positive/negative patches of the validation set indicated by the GT labels built from 3D reconstructions.

\vspace{1mm}
\noindent
\textbf{Supervision.}
As discussed in the main paper, we build the supervision based on the ground-truth depth provided by Megadepth for training. 
However, some datasets do not provide depth information.
Thus, we provide another option to build the supervision, \ie, patch-level positive and negative labels by matching the images using the SOTA dense feature matching method, RoMA~\cite{edstedt2024roma}.
The patch pairs containing more than $5$ correspondences are set as positive, and vice versa. 
As shown in Table~\ref{tab: roma}, RoMA-based supervision works comparable to the model trained with depth supervision. 
Thus, we recommend using RoMA dense correspondences as a labeling option when fine-tuning on new data. 

\begin{table}[ht]
    \centering
    \resizebox{0.75\columnwidth}{!}{
    \begin{tabular}{r| c c}
    \hline
    \tabincell{c}{Supervision} &depth-based& RoMA-based \\
    \hline
        Avg. accuracy~(\%)~$\uparrow$&\bf{74.8} &74.1\\
Avg. med. error~($^\circ$)~$\downarrow$ &\phantom{1}\bf{1.6}&\phantom{1}1.7\\
    \hline
    \end{tabular}}
     \caption{
     Average accuracy (\%) and median pose error ($^\circ$) on the retrieved top-5 images using different supervision on the test sets as the main paper.
     InLoc is excluded from the median error.}
     \label{tab: roma}
     \vspace{-0.5em}
\end{table}

\begin{table}[ht]
\centering
\vspace{-0.5em}
    \resizebox{1\columnwidth}{!}{
    \begin{tabular}{r c c c}
    \hline
Method & Backbone  & train data & pre-filter \\\hline
NetVLAD~\cite{arandjelovic2016netvlad}  & ResNet-18~\cite{resnet} & Pitts30k~\cite{torii2013visual} &-\\
CosPlace~\cite{berton2022rethinking} & ResNet-101~\cite{resnet} &SF-XL~\cite{berton2022rethinking} & -\\
CosPlace*\cite{berton2022rethinking} & ResNet-101 & MegaDepth & -\\
DINOv2~\cite{oquab2023dinov2} & ViT-G14~\cite{dosovitskiy2020image}& LVD-142M~\cite{oquab2023dinov2} &-\\
AnyLoc~\cite{keetha2023anyloc}& DINOv2 & - &-\\
SALAD~\cite{Izquierdo_CVPR_2024_SALAD}& DINOv2 &GSV-Cities~\cite{ali2022gsv} & -\\
P-NetVLAD~\cite{Hausler2021} &\multirow{2}{*}{NetVLAD}& \multirow{2}{*}{Pitts30k\& MSLS~\cite{warburg2020mapillary}}& NetVLAD\\
$\dagger$P-NetVLAD~\cite{Hausler2021}& &  & DINOv2-CLS\\
$R^2$Former~\cite{Zhu_2023_R2Former} & \multirow{2}{*}{ViT-S~\cite{dosovitskiy2020image}} & \multirow{2}{*}{MSLS~\cite{warburg2020mapillary}} & $R^2$Former\\
$\dagger$$R^2$Former~\cite{Zhu_2023_R2Former} &  &  & DINOv2-CLS\\
\bf{VOP} &DINOv2 & MegaDepth & DINOv2-CLS\\\hline
    \end{tabular}}
         \vspace{-0.5em}
    \caption{Backbones, training data, and the perfilter methods used for reranking shortlist generation are listed. }
    \label{tab:methods}
    \vspace{-0.75em}
\end{table}

\vspace{1mm}
\noindent
\textbf{Fairness Comparison.}
As shown in Table~\ref{tab:methods}, we show the backbones, training data and prefiltering methods (for reranking methods). 
We include fine-tuned results for CosPlace and results of the reranking methods tested with the same prefilter as ours.

\vspace{2mm}
\noindent
\textbf{Pose Graph construction.}
Similar to Fig.~5 shown in the main paper, here, we replace 1K random orders of queries by their predicted similarity/overlap scores. 
As shown in Fig.~\ref{fig:graph}, VOP built a larger connected component on top-1 of the queries than the competitors, also with the most number of successful RANSACs.
On the right plot (\textit{top-10}), VOP iterates more queries than $R^2$Former when ranked by the scores.
However, the run-time is comparable as VOP skipped more times of RANSAC runs.

\vspace{1mm}
\noindent
\textbf{Image Patching.}
We investigated the number of patches to be used to split the images.
The experiments were conducted on the testing scenes of MegaDepth, with all the images resized to $224^2$. 
From the trained VOP model with a patch size of $14^2$, we can extract 256 patch descriptors.
Then, average pooling is applied to aggregate the patch descriptors to different patch sizes, such as $28^2$, $56^2$, $112^2$, and $224^2$.
For example, patch size = $224^2$ will lead to a single patch of an image.
The retrieval is done on the same prefiltered image list, similarly as in the main paper. 
As shown in Table~\ref{tab:ablation_size}, the aggregated patches \eg, patch size=$224^2$, perform worse than $14^2$ on MegaDepth pose estimation and could not generalize well on Inloc localization.
This demonstrates that patch-level features can potentially improve estimated pose and other geometric problems.

\begin{table}[t]
    \centering
     \resizebox{0.95\columnwidth}{!}{\begin{tabular}{c|ccccc}
          \hline
      \multirow{2}{*}{Patch size}& AUC@10$^\circ$~$\uparrow$ & med. pose err.~$\downarrow$  &\multicolumn{2}{c}{recall@5$^\circ$, 0.5m}~$\uparrow$ \\
           & \multicolumn{2}{c}{MegaDepth} &DUC1  & DUC2\\
              \hline
         $224^2$ &67.0 &2.09& 30.8& 24.4 \\
        $112^2$ &67.0 &2.09 &38.4 & 38.9\\
       $56^2$ & 66.5 &2.17 &48.5& 57.3\\
        $28^2$ & 65.9 & 2.29 &59.1&72.5 \\
        $14^2$ &\bf{67.6} &\bf{2.03}&\bf{72.2} &\bf{77.1} \\
              \hline
    \end{tabular}}
    \caption{Ablations on different patch sizes used in inference time. We show the AUC@10$^\circ$ scores and median pose errors of the top-5 retrieved images on MegaDepth~\cite{MDLi18}, and the recall@$^\circ$, 0.5m on the top-40 of the localization data, Inloc~\cite{taira2018inloc} (DUC1, DUC2).} 
    \label{tab:ablation_size}
\end{table}

\vspace{1mm}
\noindent
\textbf{Storage \& Query Speed.}
As we reduce the dimensionality of the DINOv2 features from 1024 to 256, the embeddings of all patches of each image need a total of 512 kB, while, for AnyLoc, the storage per image is 384 kB.
While we require slightly more storage than AnyLoc~\cite{keetha2023anyloc}, the difference is small.
Compared to storing the local features in the reranking-based methods, VOP costs less.
In addition, we compare the time of querying an image from the database of different sizes using AnyLoc or VOP in seconds.
Prefiltering top 20 images by DINOv2 [CLS] token and running VOP for reranking to get top-1 out of 500 images cost 0.009 seconds, while 0.003 for AnyLoc.
Querying top-1 from 5K images (prefiltered to 100), our method costs 0.03 seconds, while AnyLoc runs in 0.02s. 
Note that VOP needs much more time for radius search without prefiltering.
We recommend using VOP as a reranking method in retrieval.}

\end{document}